\documentclass[letterpaper]{article} 
\usepackage{aaai25}  
\usepackage{times}  
\usepackage{helvet}  
\usepackage{courier}  
\usepackage[hyphens]{url}  
\usepackage{graphicx} 
\usepackage{natbib}  
\usepackage{caption} 
\usepackage{algorithm}
\usepackage{algorithmic}
\usepackage{graphicx}
\usepackage{booktabs}
\usepackage{subcaption} 
\usepackage{multirow}

\usepackage{newfloat}
\usepackage{listings}
\DeclareCaptionStyle{ruled}{labelfont=normalfont,labelsep=colon,strut=off} 
\floatstyle{ruled}
\newfloat{listing}{tb}{lst}{}
\floatname{listing}{Listing}
\title{Open-Source Protein Language Models for Function Prediction and Protein Design}
\author{
    Shivasankaran Vanaja Pandi\textsuperscript{\rm 1}\textsuperscript{\rm 2}\\
    Bharath Ramsundar\textsuperscript{\rm 1}\\
}
\affiliations{
    \textsuperscript{\rm 1}Deep Forest Sciences\\
    \textsuperscript{\rm 2}Stony Brook University\

    svanajapandi@cs.stonybrook.edu, bharath@deepforestsci.com , 
}

\usepackage{bibentry}

\begin{document}

\maketitle

\begin{abstract}

Protein language models (PLMs) have shown promise in improving the understanding of protein sequences, contributing to advances in areas such as function prediction and protein engineering. However, training these models from scratch requires significant computational resources, limiting their accessibility. To address this, we integrate a PLM into DeepChem, an open-source framework for computational biology and chemistry, to provide a more accessible platform for protein-related tasks.

We evaluate the performance of the integrated model on various protein prediction tasks, showing that it achieves reasonable results across benchmarks. Additionally, we present an exploration of generating plastic-degrading enzyme candidates using the model's embeddings and latent space manipulation techniques. While the results suggest that further refinement is needed, this approach provides a foundation for future work in enzyme design. This study aims to facilitate the use of PLMs in research fields like synthetic biology and environmental sustainability, even for those with limited computational resources.

\end{abstract}

\section{Introduction}

Protein language models (PLMs) have shown promise in improving the understanding of protein sequences by learning from large datasets \cite{elnaggar2021prottrans}\cite{rao2020transformer}\cite{lin2023evolutionary}\cite{meier2021language}. These models are typically trained on vast collections of protein sequences, using significant computational resources. While the scale of these models enables them to capture complex biological patterns, the computational burden associated with training such models from scratch makes it an impractical task for many researchers, especially those without substantial infrastructure.

For many researchers, particularly in fields like biology and chemistry, access to pretrained PLMs is essential for advancing their work. However, the technical barrier remains high. The ability to use these pretrained models without needing extensive computational infrastructure or the expertise to fine-tune them would be a valuable resource. Pretrained models can be directly applied or adapted to specific downstream tasks, simplifying their usage and reducing the effort required for implementation.

Moreover, many potential users of these PLMs are biologists and chemists who may not have extensive backgrounds in computer science or machine learning. To address this challenge, we integrate a PLM into DeepChem \cite{ramsundar2019deep}, an open-source machine learning framework widely used by biologists and chemists. This integration offers an accessible method for leveraging advanced protein sequence models within a familiar framework, removing the need for users to engage with the complexities of model training or fine-tuning.

\begin{figure*}[t]
\centering
\includegraphics[width=0.8\linewidth]{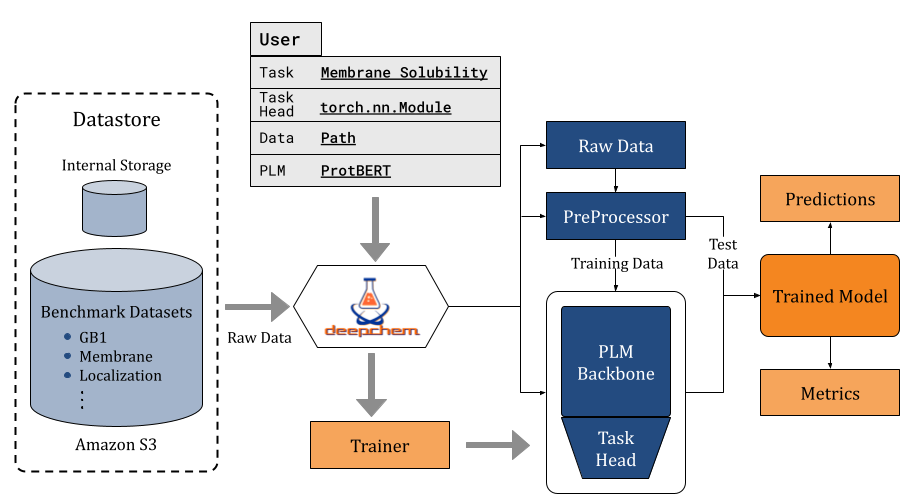} 
\caption{DeepChem Pipeline Illustration}
\label{fig2}
\end{figure*}

Our contributions are as follows:
\begin{itemize}
\item We evaluate the performance of the pretrained PLM on standard protein-related benchmarks within DeepChem, showing its applicability to protein-related tasks.
\item We present an exploration of the model's practical application to a specific bioinformatics problem, illustrating its potential for real-world tasks.
\item We open-source our implementation, providing the broader scientific community with easy access to the capabilities of large PLMs without requiring extensive computational resources or advanced expertise.
\end{itemize}

Through this integration, we aim to lower the barriers to using large-scale PLMs in scientific research, empowering a wider range of researchers to harness the potential of these models for their work.


\section{Related Works}

The release of extensive protein datasets, such as UniProt \cite{uniprot2019uniprot}, BFD\cite{jumper2021highly} has greatly advanced the development and application of large language models (LLMs) in analyzing protein sequences. By representing protein sequences as "biological text", PLMs can capture subtle structural, evolutionary, and functional patterns that traditional bioinformatics methods may struggle to identify\cite{lin2022language}\cite{lin2023evolutionary}. This capacity for nuanced understanding has enabled PLMs to excel in various biological tasks, including protein function prediction, secondary structure modeling, and enzyme engineering\cite{elnaggar2021prottrans}\cite{rao2019evaluating}\cite{rao2020transformer}. These applications are crucial in fields like drug discovery, molecular biology, and synthetic biology, where understanding protein behavior and interactions is paramount.

In protein design, PLMs are proving to be powerful tools for generating novel proteins with tailored properties, a task typically requiring substantial experimental effort\cite{madani2020progen}\cite{madani2023large}. PLMs can model the relationships between sequence and function, enabling the generation of protein variants that retain or even enhance desired characteristics. For example, Madani et al. \cite{madani2020progen} introduced a transformer-based conditional generation model that produces protein sequences conditioned on specific attributes or “control tags,” such as structural or functional classifications derived from the UniProt database. This approach allows researchers to generate proteins with targeted properties or functionalities, advancing efforts in areas such as enzyme engineering, where novel enzymes are designed for tasks like bioremediation, synthetic pathway creation, or industrial catalysis. Such conditional generation techniques illustrate the transformative potential of PLMs in protein engineering, providing a new avenue for designing proteins with applications in environmental sustainability and biomedical research.

    

\section{Experiments}
\subsection{Benchmarking}

To assess the performance of our ProtBERT integration within the DeepChem framework, we conducted a comprehensive benchmarking study. Following the training setup specified by the original ProtBERT authors, we fine-tuned ProtBERT on a range of protein-specific downstream tasks, utilizing DeepChem’s built-in datasets and evaluation utilities. Our objective was to evaluate both the functionality and performance of ProtBERT in a reproducible pipeline, leveraging DeepChem’s robust data processing and model evaluation tools.

Following the original authors' methodology, ProtBERT was initially pretrained on the UniRef dataset, a comprehensive collection of protein sequences, to learn general protein representations. However, unlike the original ProtBERT, which was trained on 216 million protein sequences, our pretraining was limited to 1 million sequences. After pretraining, ProtBERT was fine-tuned on task-specific datasets for each benchmark task. The training process was carried out on a single A100 GPU instance in Google Colaboratory over a span of four days.

\subsubsection{Benchmark Tasks}
We evaluated ProtBERT on a comprehensive set of protein classification and regression tasks within DeepChem, including sub-cellular localization, membrane solubility, epitope region, and GB1 mutational landscape prediction. For every task, we fine-tuned a one-layer MLP network on top of ProtBERT's embeddings. For each classification task, we recorded the accuracy, while for the regression tasks, we reported Spearman's rank correlation coefficient ($\rho$). These benchmarks provide insights into ProtBERT's ability to address diverse protein-related problems, further establishing a reproducible baseline for future protein language models integrated with DeepChem. The raw data for each task was processed into files compatible with the DeepChem framework, and training was conducted on an A100 GPU instance in Google Colaboratory.

\paragraph{Sub-cellular Localization Prediction} 
This classification task involves predicting the cellular compartments where proteins are likely to be located. Accurate sub-cellular localization \cite{deeploc} predictions are crucial for understanding protein functions in biological contexts. In this work, we compare our results against the original metrics published by the ProtBERT\cite{elnaggar2021prottrans} authors, as we aim to replicate these results within our DeepChem implementation.

\paragraph{Membrane Solubility Classification}
Membrane solubility prediction \cite{deeploc} is a binary classification task where the model determines whether a protein is soluble or membrane-bound. This task is critical for drug discovery and protein engineering, as membrane proteins play essential roles in cellular signaling and transport. In this work, we compare our results against the original metrics published by the ProtBERT \cite{elnaggar2021prottrans} authors, as we aim to replicate these results with our implementation.

\paragraph{Epitope Region Prediction}
Predicting epitope regions, or immunologically active segments of proteins, is a challenging task due to the complexity of protein-antibody interactions \cite{hou2017seeing}. In this classification task, ProtBERT is used to identify likely epitope regions within protein sequences. In our evaluation, we compare against the BERT model specifically trained for this task and published by the authors of the ProteinGLUE Benchmark\cite{capel2022proteinglue}.

\paragraph{GB1 Mutational Landscape Prediction}
GB1 mutational landscape prediction is a regression task that aims to predict the activity of protein variants with multiple mutations, focusing on epistatic interactions between mutations\cite{dallago2021flip}. These interactions can complicate predictions, as the effect of one mutation depends on others. The task involves learning from variants with fewer mutations to predict the activity of more complex variants, with results reported using Spearman’s rank correlation coefficient ($\rho$) to capture the model’s ability to predict mutational effects.

These benchmarks establish a reproducible baseline for future protein language models integrated with DeepChem, showing that ProtBERT performs well across diverse tasks relevant to protein function and property prediction. By demonstrating strong performance in both classification and regression contexts, ProtBERT underscores the potential of protein language models in advancing computational biology research.

\subsection{Generating Plastic-Degrading Enzymes}

In this case study, we explore the generation of plastic-degrading protein sequences using a variational autoencoder (VAE) approach, integrated with a pretrained ProtBERT model. To generate protein sequences that may resemble known plastic-degrading enzymes, we employ a seed-based generation technique. This involves encoding a set of known plastic-degrading proteins into latent representations and adding controlled noise to these representations, which introduces variability in the generated protein sequences.

\begin{figure}[t] 
\centering
\begin{subfigure}[b]{\columnwidth} 
    \centering
    \includegraphics[width=0.9\linewidth]{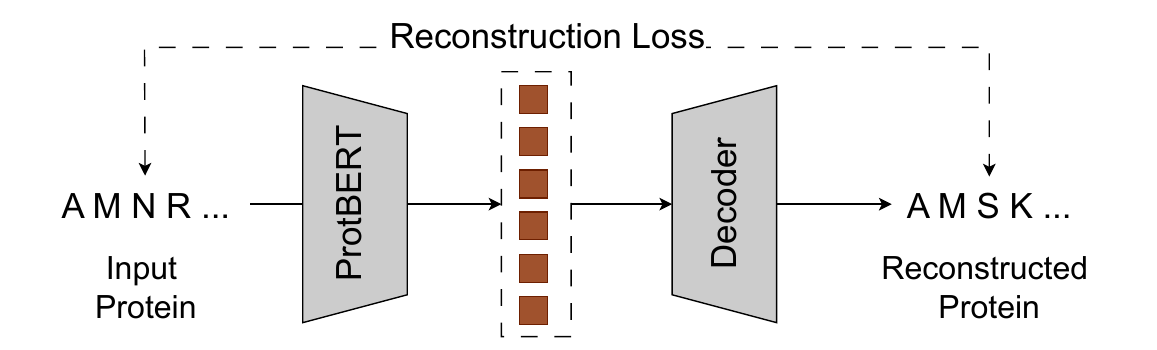} 
    \label{subfig1}
\end{subfigure}

\vspace{0.5cm} 

\begin{subfigure}[b]{\columnwidth} 
    \centering
    \includegraphics[width=0.9\linewidth]{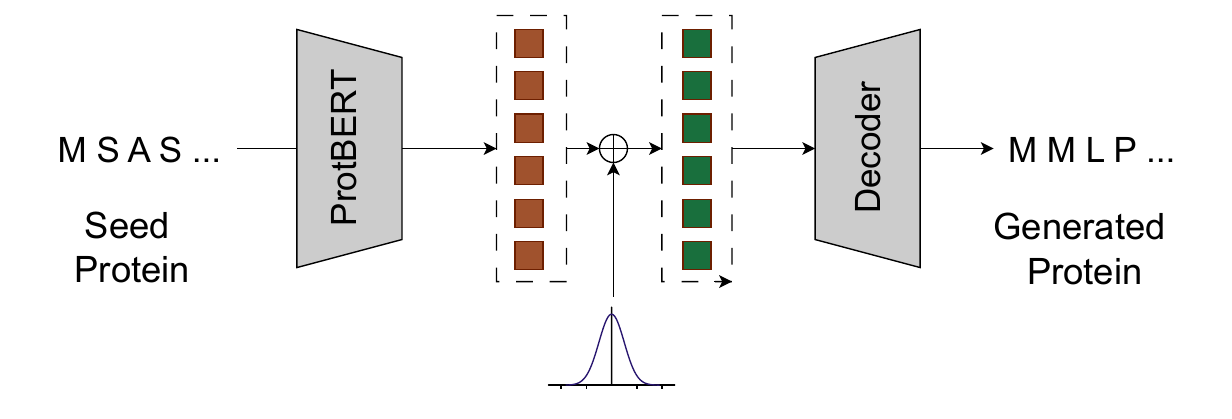} 
    \label{subfig2}
\end{subfigure}

\caption{Protein generation pipeline: (a) Overview of the training pipeline for the protein generation model (b) Protein generation workflow demonstrating the use of a seed protein sequence to generate novel proteins with targeted properties.}
\label{fig:main}
\end{figure}

\begin{figure}[h] 
\centering
\begin{subfigure}[b]{\columnwidth} 
    \centering
    \includegraphics[width=0.62\linewidth]{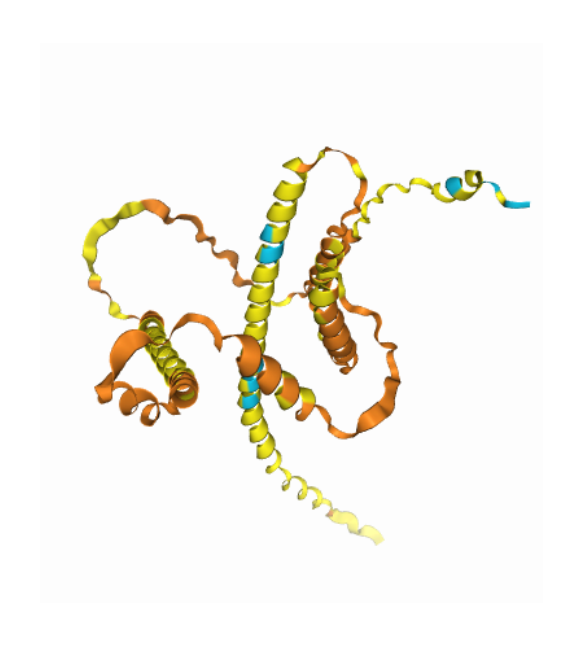} 
    \caption{PlDDT score: 55.01. TM-score: 0.363 with V9SDT4}
    \label{subfig1}
\end{subfigure}


\begin{subfigure}[b]{\columnwidth} 
    \centering
    \includegraphics[width=0.62\linewidth]{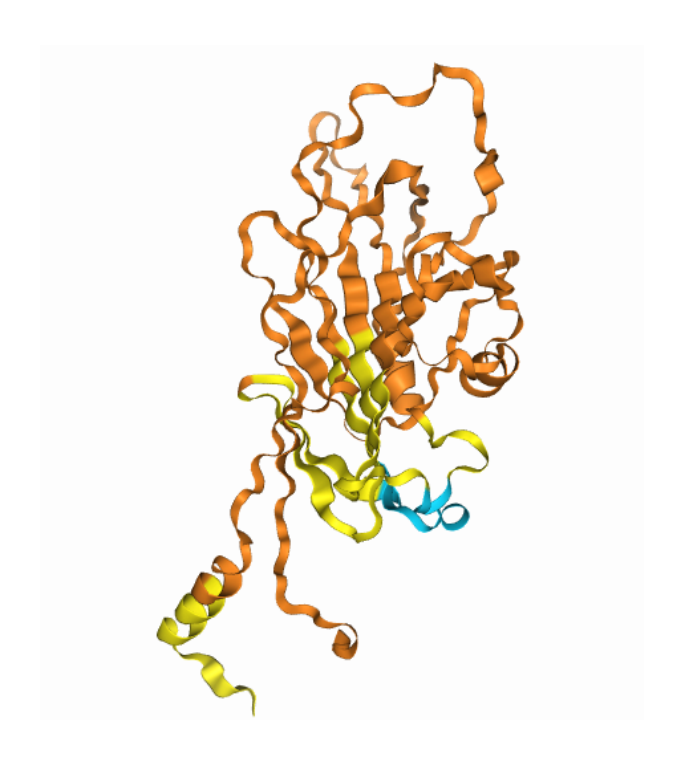} 
    \caption{PlDDT score: 63.22. TM-score: 0.367 with F2WL50}
    \label{subfig2}
\end{subfigure}

\caption{Qualitative Results of Generated Proteins. Two proteins generated by our method. The protein structures are color-coded according to the plDDT\cite{jumper2021highly} score: Blue: very high (plDDT $>$ 90), teal: confident (90 $>$ plDDT $>$ 70), yellow: low (70 $>$ plDDT $>$ 50), orange: very low (plDDT $<$ 50) }
\label{fig:qual_res}
\end{figure}


\subsubsection{Training the Model}

The decoder model was trained on a dataset of 100,000 protein sequences sourced from the UniRef dataset, filtered to include sequences with lengths below 512 tokens. Training was conducted for 5 epochs on an A100 GPU instance in Google Colaboratory, utilizing the pretrained ProtBERT model to generate high-quality embeddings and sequence representations. This pretraining step proved instrumental in learning latent representations that could be effectively leveraged for protein generation tasks.

\subsubsection{Latent Space Manipulation for Controlled Generation}

To generate protein sequences that are similar to plastic-degrading enzymes, we employed a targeted approach by manipulating the latent representation space around known plastic-degrading proteins. Rather than randomly sampling from the latent space, we encoded one of the known plastic-degrading proteins as a `seed` latent representation. This approach directed the generation process toward proteins with similar properties.

To introduce diversity and prevent generating excessively similar sequences to the input seed, we added Gaussian noise to the seed's latent representation. By controlling the noise level, we influenced the degree of variation from the original seed protein, with higher noise levels leading to more distinct sequences. This process allowed us to explore variations within the protein sequence space that could potentially resemble plastic-degrading enzymes, while still maintaining control over the extent of variation from the known proteins.

\section{Results}

\subsection{Benchmarking}

Our results demonstrate that ProtBERT, as integrated into DeepChem, achieves competitive performance across various benchmarks, despite being pretrained on only 1M protein sequences. Notably, ProtBERT showed strong performance in classification tasks, such as sub-cellular localization, membrane solubility, and Epitope Region Prediction. Furthermore, for the regression tasks of GB1 Fitness Prediction, ProtBERT demonstrated high Spearman’s $\rho$, reflecting its potential in capturing complex protein sequence relationships for regression-based problems.

Table \ref{tab-metrics} presents the specific metrics for each benchmarking task, further showcasing ProtBERT’s performance across these tasks.

The integration of ProtBERT within DeepChem also enabled a streamlined evaluation process, making ProtBERT a powerful addition to the DeepChem toolkit for a wide range of protein-related machine learning tasks.

These findings validate the compatibility and effectiveness of ProtBERT within the DeepChem framework, paving the way for future integrations of similar large-scale protein language models.
\vspace{-0.3cm}
\subsection{Generating Plastic-Degrading Enzymes}

To assess the effectiveness of our model in generating plausible plastic-degrading enzymes, we utilized AlphaFold for structural prediction. By estimating the 3D structures of the generated protein sequences, we aimed to verify whether the generated enzymes exhibit structural features commonly found in naturally occurring plastic-degrading enzymes. This structural evaluation serves as a proxy for assessing the biological relevance of the generated proteins, ensuring they have realistic conformations, secondary structures, and potential active sites.

Our approach involved generating multiple protein sequences based on latent representations of known plastic-degrading enzymes, with added Gaussian noise to introduce variation while staying within the desired functional space. Each generated sequence was then passed through AlphaFold for structural estimation. We compared the structural predictions to identify proteins that shared similarities with known plastic-degrading enzymes. Figure \ref{fig:qual_res} presents examples of the generated proteins, which showed average pLLDT values greater than 50, suggesting potential for plastic degradation activity.

\begin{table}[t]
    \centering
    \renewcommand{\arraystretch}{1.5} 
    \begin{tabular}{|c|c|c|}
        \hline
        \multirow{2}{*}{\textbf{Task}} & \multicolumn{2}{c|}{\textbf{Performance}} \\
        \cline{2-3}
        & \textbf{$PB_{orig}$} & \textbf{$PB_{DC}$} \\
        \hline
        Sub-cellular Localization & 74 & 69.7 \\
        Membrane Solubility  & 89 & 85.2 \\
        Epitope Region Prediction & 69.51 & 66.73 \\
        GB1 Fitness Prediction  & 0.63 & 0.43 \\
        \hline
    \end{tabular}
    \caption{Performance comparison of protein language models on various protein prediction tasks. $PB_{orig}$ refers to the original ProtBERT model, trained on 216 million protein sequences, while $PB_{DC}$ represents the ProtBERT model integrated into DeepChem, trained on 1 million protein sequences. Performance is reported for sub-cellular localization (accuracy), membrane solubility (accuracy), epitope region prediction (AUC ROC), and GB1 fitness prediction (Spearman correlation).}

    \label{tab-metrics}
\end{table}

\vspace{-0.2cm} 
\section{Conclusion}
In this work, we introduced the integration of ProtBERT, a protein language model, into the DeepChem library, making it more accessible for protein-related tasks in computational biology. Through an evaluation, our integration achieved results that are comparable to existing models on a range of protein classification and regression tasks.

Our case study on generating plastic-degrading enzymes highlights the potential applications of ProtBERT's latent space. By using embeddings of known plastic-degrading proteins and introducing controlled noise, we generated novel enzyme candidates, which were structurally validated with AlphaFold. While the results are promising, further investigation is needed to confirm the functional relevance of these candidates. Future studies could leverage Quantum Mechanics/Molecular Mechanics (QM/MM) simulations\cite{kurniawan2022protein} to further validate the generated proteins, offering a complementary approach to assess their enzymatic activity and stability.

The integration of ProtBERT into DeepChem enhances its utility for protein-related research and provides a platform for reproducible and accessible model development within the scientific community. We hope this work serves as a step forward in the use of protein language models, supporting future efforts in protein engineering, drug discovery, and environmental biotechnology.

\bibliography{aaai25}

\begin{thebibliography}{16}
\providecommand{\natexlab}[1]{#1}

\bibitem[{Almagro~Armenteros et~al.(2017)Almagro~Armenteros, S{\o}nderby, S{\o}nderby, Nielsen, and Winther}]{deeploc}
Almagro~Armenteros, J.~J.; S{\o}nderby, C.~K.; S{\o}nderby, S.~K.; Nielsen, H.; and Winther, O. 2017.
\newblock DeepLoc: prediction of protein subcellular localization using deep learning.
\newblock \emph{Bioinformatics}, 33(21): 3387--3395.

\bibitem[{Capel et~al.(2022)Capel, Weiler, Dijkstra, Vleugels, Bloem, and Feenstra}]{capel2022proteinglue}
Capel, H.; Weiler, R.; Dijkstra, M.; Vleugels, R.; Bloem, P.; and Feenstra, K.~A. 2022.
\newblock ProteinGLUE multi-task benchmark suite for self-supervised protein modeling.
\newblock \emph{Scientific Reports}, 12(1): 16047.

\bibitem[{Consortium(2019)}]{uniprot2019uniprot}
Consortium, U. 2019.
\newblock UniProt: a worldwide hub of protein knowledge.
\newblock \emph{Nucleic acids research}, 47(D1): D506--D515.

\bibitem[{Dallago et~al.(2021)Dallago, Mou, Johnston, Wittmann, Bhattacharya, Goldman, Madani, and Yang}]{dallago2021flip}
Dallago, C.; Mou, J.; Johnston, K.~E.; Wittmann, B.~J.; Bhattacharya, N.; Goldman, S.; Madani, A.; and Yang, K.~K. 2021.
\newblock FLIP: Benchmark tasks in fitness landscape inference for proteins.
\newblock \emph{bioRxiv}, 2021--11.

\bibitem[{Elnaggar et~al.(2021)Elnaggar, Heinzinger, Dallago, Rehawi, Wang, Jones, Gibbs, Feher, Angerer, Steinegger et~al.}]{elnaggar2021prottrans}
Elnaggar, A.; Heinzinger, M.; Dallago, C.; Rehawi, G.; Wang, Y.; Jones, L.; Gibbs, T.; Feher, T.; Angerer, C.; Steinegger, M.; et~al. 2021.
\newblock Prottrans: Toward understanding the language of life through self-supervised learning.
\newblock \emph{IEEE transactions on pattern analysis and machine intelligence}, 44(10): 7112--7127.

\bibitem[{Hou et~al.(2017)Hou, De~Geest, Vranken, Heringa, and Feenstra}]{hou2017seeing}
Hou, Q.; De~Geest, P.~F.; Vranken, W.~F.; Heringa, J.; and Feenstra, K.~A. 2017.
\newblock Seeing the trees through the forest: sequence-based homo-and heteromeric protein-protein interaction sites prediction using random forest.
\newblock \emph{Bioinformatics}, 33(10): 1479--1487.

\bibitem[{Jumper et~al.(2021)Jumper, Evans, Pritzel, Green, Figurnov, Ronneberger, Tunyasuvunakool, Bates, {\v{Z}}{\'\i}dek, Potapenko et~al.}]{jumper2021highly}
Jumper, J.; Evans, R.; Pritzel, A.; Green, T.; Figurnov, M.; Ronneberger, O.; Tunyasuvunakool, K.; Bates, R.; {\v{Z}}{\'\i}dek, A.; Potapenko, A.; et~al. 2021.
\newblock Highly accurate protein structure prediction with AlphaFold.
\newblock \emph{nature}, 596(7873): 583--589.

\bibitem[{Kurniawan and Ishida(2022)}]{kurniawan2022protein}
Kurniawan, J.; and Ishida, T. 2022.
\newblock Protein model quality estimation using molecular dynamics simulation.
\newblock \emph{ACS omega}, 7(28): 24274--24281.

\bibitem[{Lin et~al.(2022)Lin, Akin, Rao, Hie, Zhu, Lu, dos Santos~Costa, Fazel-Zarandi, Sercu, Candido et~al.}]{lin2022language}
Lin, Z.; Akin, H.; Rao, R.; Hie, B.; Zhu, Z.; Lu, W.; dos Santos~Costa, A.; Fazel-Zarandi, M.; Sercu, T.; Candido, S.; et~al. 2022.
\newblock Language models of protein sequences at the scale of evolution enable accurate structure prediction.
\newblock \emph{BioRxiv}, 2022: 500902.

\bibitem[{Lin et~al.(2023)Lin, Akin, Rao, Hie, Zhu, Lu, Smetanin, Verkuil, Kabeli, Shmueli et~al.}]{lin2023evolutionary}
Lin, Z.; Akin, H.; Rao, R.; Hie, B.; Zhu, Z.; Lu, W.; Smetanin, N.; Verkuil, R.; Kabeli, O.; Shmueli, Y.; et~al. 2023.
\newblock Evolutionary-scale prediction of atomic-level protein structure with a language model.
\newblock \emph{Science}, 379(6637): 1123--1130.

\bibitem[{Madani et~al.(2023)Madani, Krause, Greene, Subramanian, Mohr, Holton, Olmos, Xiong, Sun, Socher et~al.}]{madani2023large}
Madani, A.; Krause, B.; Greene, E.~R.; Subramanian, S.; Mohr, B.~P.; Holton, J.~M.; Olmos, J.~L.; Xiong, C.; Sun, Z.~Z.; Socher, R.; et~al. 2023.
\newblock Large language models generate functional protein sequences across diverse families.
\newblock \emph{Nature Biotechnology}, 41(8): 1099--1106.

\bibitem[{Madani et~al.(2020)Madani, McCann, Naik, Keskar, Anand, Eguchi, Huang, and Socher}]{madani2020progen}
Madani, A.; McCann, B.; Naik, N.; Keskar, N.~S.; Anand, N.; Eguchi, R.~R.; Huang, P.-S.; and Socher, R. 2020.
\newblock Progen: Language modeling for protein generation.
\newblock \emph{arXiv preprint arXiv:2004.03497}.

\bibitem[{Meier et~al.(2021)Meier, Rao, Verkuil, Liu, Sercu, and Rives}]{meier2021language}
Meier, J.; Rao, R.; Verkuil, R.; Liu, J.; Sercu, T.; and Rives, A. 2021.
\newblock Language models enable zero-shot prediction of the effects of mutations on protein function.
\newblock \emph{Advances in neural information processing systems}, 34: 29287--29303.

\bibitem[{Ramsundar et~al.(2019)Ramsundar, Eastman, Walters, and Pande}]{ramsundar2019deep}
Ramsundar, B.; Eastman, P.; Walters, P.; and Pande, V. 2019.
\newblock \emph{Deep learning for the life sciences: applying deep learning to genomics, microscopy, drug discovery, and more}.
\newblock " O'Reilly Media, Inc.".

\bibitem[{Rao et~al.(2019)Rao, Bhattacharya, Thomas, Duan, Chen, Canny, Abbeel, and Song}]{rao2019evaluating}
Rao, R.; Bhattacharya, N.; Thomas, N.; Duan, Y.; Chen, P.; Canny, J.; Abbeel, P.; and Song, Y. 2019.
\newblock Evaluating protein transfer learning with TAPE.
\newblock \emph{Advances in neural information processing systems}, 32.

\bibitem[{Rao et~al.(2020)Rao, Meier, Sercu, Ovchinnikov, and Rives}]{rao2020transformer}
Rao, R.; Meier, J.; Sercu, T.; Ovchinnikov, S.; and Rives, A. 2020.
\newblock Transformer protein language models are unsupervised structure learners.
\newblock \emph{Biorxiv}, 2020--12.

\end{thebibliography}

\end{document}